\documentclass[runningheads]{llncs}
\usepackage{graphicx}
\usepackage{array}
\usepackage{booktabs} 
\usepackage{amsmath}
\usepackage{amsfonts}
\usepackage{amssymb}

\begin{document}
\title{Instruct Large Language Models to Generate Scientific Literature Survey Step by Step}
\titlerunning{Instruct LLMs to Generate Scientific Literature Survey Step by Step}
\author{Yuxuan Lai\inst{1,2} \and
Yupeng Wu\inst{3} \and
Yidan Wang\inst{1} \and Wenpeng Hu\inst{4} \and Chen Zheng\inst{1,2} }
\authorrunning{Y. Lai et al.}
\institute{The Open University of China, Beijing, China \\
\email{erutan@pku.org.cn, laiyx@ouchn.edu.cn}\\
\and
Engineering Research Center of Integration and Application of Digital Learning Technology, Ministry of Education, Beijing, China
\and
FloatMiracle, Beijing, China
\and
PLA Academy of Military Science}
\maketitle              
\begin{abstract}
Automatically generating scientific literature surveys is a valuable task that can significantly enhance research efficiency. 
However, the diverse and complex nature of information within a literature survey poses substantial challenges for generative models. 
In this paper, we design a series of prompts to systematically leverage large language models (LLMs), enabling the creation of comprehensive literature surveys through a step-by-step approach.
Specifically, we design prompts to guide LLMs to sequentially generate the title, abstract, hierarchical headings, and the main content of the literature survey. 
We argue that this design enables the generation of the headings from a high-level perspective. 
During the content generation process, this design effectively harnesses relevant information while minimizing costs by restricting the length of both input and output content in LLM queries.
Our implementation with Qwen-long achieved third place in the NLPCC 2024 \textit{Scientific Literature Survey Generation} evaluation task, with an overall score only 0.03\% lower than the second-place team. 
Additionally, our soft heading recall is 95.84\%, the second best among the submissions. 
Thanks to the efficient prompt design and the low cost of the Qwen-long API, our method reduces the expense for generating each literature survey to 0.1 RMB, enhancing the practical value of our method.
\end{abstract}
\section{Introduction}

Conducting a literature survey is a crucial component of scientific research\cite{fink2019conducting}. 
However, as the volume of scientific publications continues to grow exponentially and the rate of new publications accelerates\cite{liang2024mapping}, it costs the researchers increasing amounts of time on this task. 
With the advanced capabilities of large language models (LLMs)\cite{openai2023gpt4,ouyang2022training,bai2023qwen}, AI-generated literature surveys offer a promising solution by significantly reducing the time and effort required. 

However, generating scientific literature surveys presents several challenges. Firstly, these surveys are typically very lengthy, and even advanced models like GPT-4\cite{openai2023gpt4} and Claude 3\cite{anthropic2024claude} are constrained by a 4k-8k token output limit, making it impossible to generate a survey in a single pass. Attempting to generate the entire content sequentially based on all previous content can also be problematic, as it may cause the LLM to forget earlier information or fail to maintain a consistent structure without a comprehensive plan.
Secondly, scientific literature surveys cover a vast amount of information. Transmitting this information to LLMs and generating a coherent survey can be very expensive in terms of LLM API usage, which diminishes its practical value.

\begin{figure}[t!]
    \centering
    \resizebox{0.92\textwidth}{!}{
    \includegraphics[width=1\textwidth]{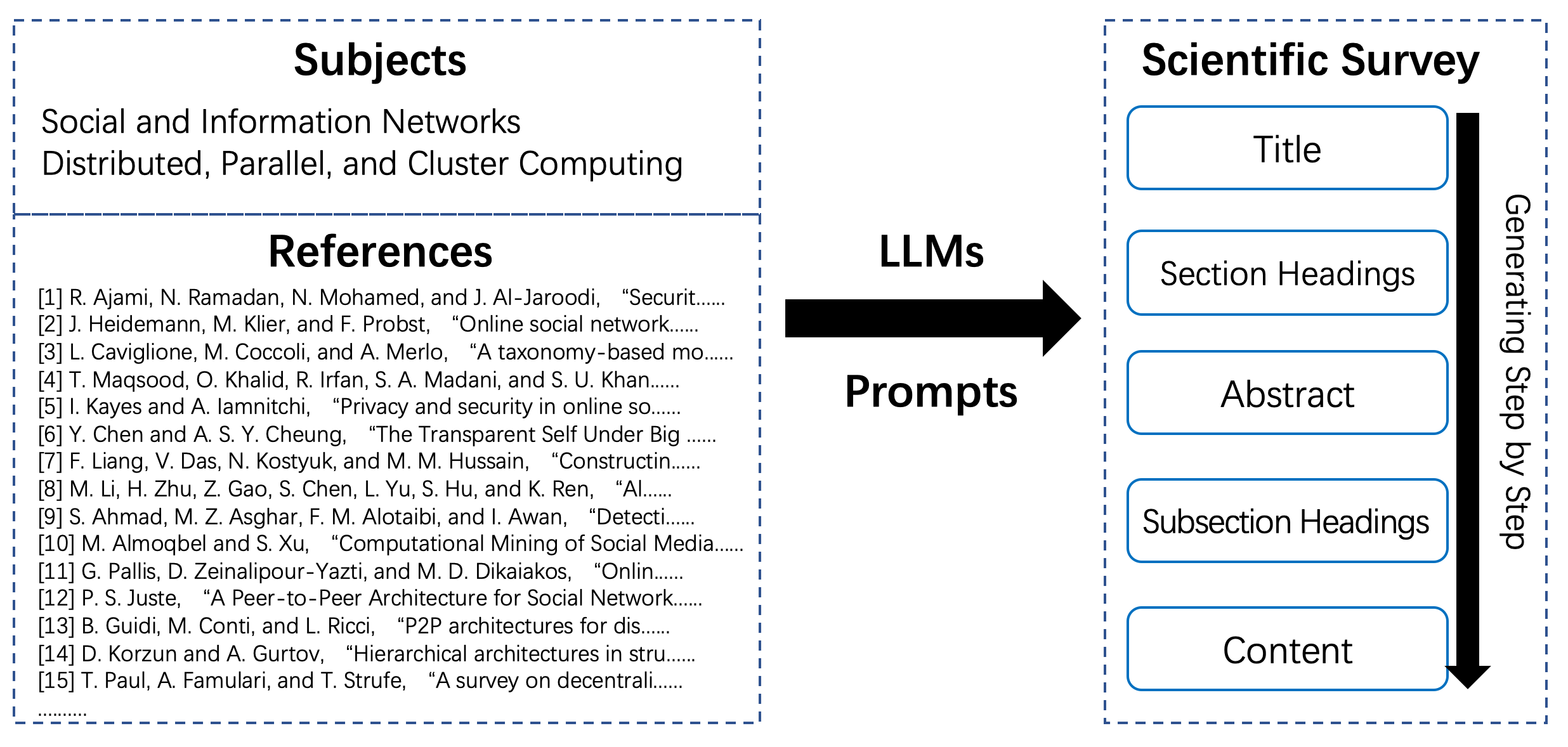}
    }
    \caption{An illustration of the scientific survey generation task, and our step-by-step generation approach.}
    \label{figure1}
\end{figure}

In this paper, we propose a novel approach to scientific literature survey generation that leverages LLMs in a step-by-step prompting manner. 
As shown in Fig.~\ref{figure1}, given the subjects and several reference papers, we design a series of prompts that guide LLMs to sequentially generate the title, abstract, hierarchical headings, and the main content of a literature survey. 
By breaking down the task into manageable steps, LLMs can maintain a high-level perspective while generating headings, thereby improving the coherence and relevance of the generated surveys. 
Additionally, after generating the headings, the main content generation can be conditioned on this structure rather than on all previous content, significantly reducing the cost of API usage.

We implemented our system using the Qwen-long version of Alibaba's Tongyi Qianwen model\cite{bai2023qwen}. Our system was submitted in the NLPCC 2024 \textit{Scientific Literature Survey Generation} task, where it achieved third place, with an overall score of 61.11, just 0.03\% behind the second-place team.  
Additionally, our method demonstrated a soft heading recall of 95.84\%, the second highest among all submissions and only 1.17\% lower than the first-place team. 
These results highlight the effectiveness of our approach in maintaining logical and coherent survey structures, which can be attributed to the plan-then-generate strategy employed in our sequential generation process.
Thanks to the efficient prompt design and the low cost of the Qwen-long API, our method reduces the expense of generating each literature survey to 0.1 RMB, enhancing its practical value.

The remainder of this paper is organized as follows: 
Section 2 reviews related works. 
Section 3 details our prompt design methodology and the step-by-step generation process. 
Section 4 presents the experimental setup and evaluation results, as well as discussing the advantages and limitations of our method. 
Finally, Section 5 concludes the paper. 

\section{Related Works}

\textbf{Automatic literature survey generation} has garnered increasing attention in recent years due to its potential to enhance research efficiency and reduce the time required for literature review processes. 
However, because of the inherent challenges involved, few works have provided a systematic methodology to leverage AI tools to address this task comprehensively. 
Recently, AutoSurvey \cite{wang2024autosurvey}, a contemporary work to this paper, proposes equipping LLMs with initial retrieval and outline generation capabilities to automatically generate literature surveys. 
Additionally, some surveys \cite{schulhoff2024prompt} are claimed to be written with the help of LLMs.

In this paper, we focus on the NLPCC 2024 \textit{Scientific Literature Survey Generation} evaluation task \cite{tian2024overview}. 
To the best of our knowledge, this is the first public evaluation task on automatic literature survey generation. 
Our system participated in this contest and achieved third place, with an overall score only 0.03\% lower than the second-place team.

\textbf{Leveraging LLMs for generating long-form text} has become a rapidly evolving area of research. 
Some efforts have explored novel attention mechanisms to model long contexts effectively \cite{dao2022flashattention,ding2023longnet,xiaoefficient}. 
However, recent studies \cite{li2023loogle,li2024long} point out that these long-context LLMs cannot fully utilize the information embedded in the extensive context window.

Instead of extending the context window of LLMs, some works explore the use of outlines to organize long text generation. 
\cite{yang2022doc} employ detailed outlines to improve the coherence of long story generation. 
\cite{lee2024navigating} propose a pipeline that includes outline generation, information retrieval augmentation, augmented outline generation, and content generation for long blog creation. 
In this paper, we also adopt an outline-based generation strategy to manage the long context required for generating scientific literature surveys in a step-by-step manner.

\section{Methodology}

\subsection{Task Formulation}
\label{sec:task}

In this paper, we address the NLPCC 2024 \textit{Scientific Literature Survey Generation} evaluation task, which focuses on generating literature surveys based on given subjects and reference papers.

Specifically, illustrated in Fig.~\ref{figure1}, given a list of reference papers \( R = \{r_1, r_2, ..., r_n\} \) and a list of key subjects (topics) \( S = \{s_1, s_2, ..., s_m\} \), our goal is to produce a well-structured literature survey \( \left<T, A, M \right>\). The output includes the title $T$, abstract $A$, and the main body $M=\left\{(h_1, c_1), (h_2, c_2), ..., (h_k, c_k) \right\}$, where $h_i$ represents the headings at different levels and $c_i$ denotes the corresponding contents.

Notice that we did not utilize the \textit{reference content} field provided in the test set because the NLPCC 2024 task guideline claimed that only the \textit{subject} and \textit{reference} were provided for testing. 
Additionally, we did not employ any retrieval-based methods to augment the reference content based on the titles of the reference papers, as the guidelines explicitly state that \textit{data other than the training set cannot be used in the process of model development}.

\begin{figure}[t!]
    \centering
    \resizebox{0.96\textwidth}{!}{
    \includegraphics[width=1\textwidth]{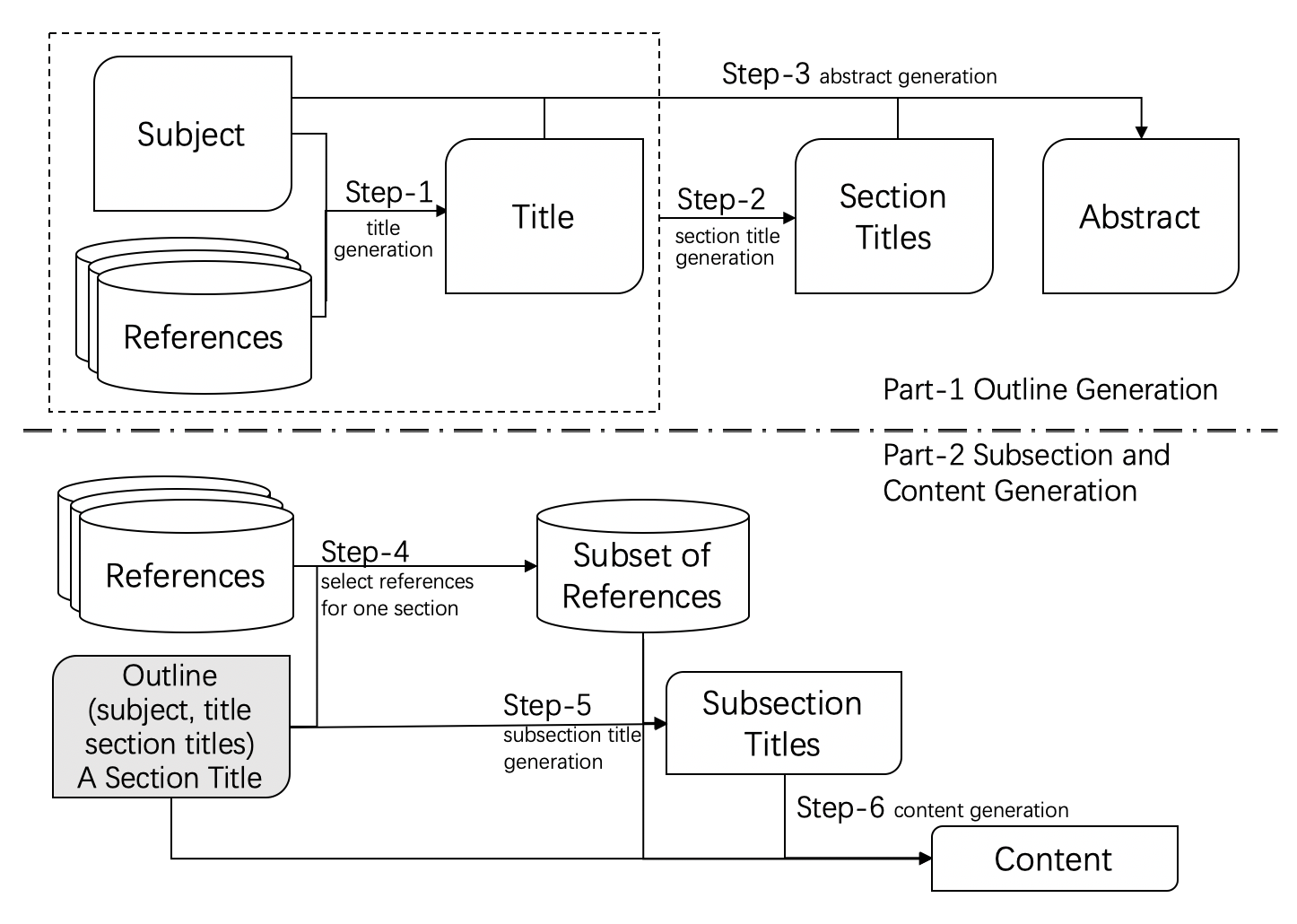}
    }
    \caption{An illustration of our step-by-step literature survey generation framework.}
    \label{figure2}
\end{figure}

\subsection{Overview}

The framework for our step-by-step literature survey generation is illustrated in Fig.~\ref{figure2}. The generation pipeline consists of six steps, divided into two main phases.

In the first phase, {outline generation}, we instruct the LLM to sequentially generate the title, section headings, and abstract. Each step is conditioned on all previous outputs to ensure a coherent planning of the entire literature survey.

In the second phase, {subsection and content generation}, we select references for each section and then generate subsection headings and content based on the generated outline and the selected references. This design ensures that the generation of each subsection is aware of the overall structure of the paper, while also reducing the input length to save costs on LLM API usage.

\subsection{Prompt Design}

In this section, we will elaborate on the prompt designs for each of the six steps in Fig.~\ref{figure2}.

\paragraph{Step 1: Title Generation} - 
We instruct the LLM to generate a suitable title for the literature survey based on the provided subject and reference papers using the following prompt. We use semicolons (;) and newline characters (\textbackslash n) to concatenate multiple subjects and references, respectively.

\begin{table}[h!]
\small
\centering
\begin{tabular}{p{11.7cm}}
\textbf{Prompt-1, title generation:}\\
There is an academic puzzle for you. \\
I will give you a list of references. In subject ``\textit{\{subject\}}", a survey paper existing with these references. You will give me a guess of the title of the survey paper.\\
Here are the references: \textit{\{references\}}\\
The output should be in several lines, and the content in the last one is your answer (the title that you guess).\\
Only one guess is required. The title should start with ``Title: ".
\end{tabular}
\end{table}

We observe that a chain-of-thought strategy \cite{wei2022chain} is generally integrated into existing LLM products, causing the model likely to generate explanations as well. We do not penalize the generation of additional information. Instead, we instruct the model to generate the title on the last line with a specific marker, \textit{Title:}. 
We verify the output format, and if the specific marker is not found, we reinstruct the LLM while retaining the memory. The same prompt is used, but we add a prefix: \textit{The response format is incorrect. Note that only one guess is required. The title should start with ``Title: "}.

\paragraph{Step 2: Section Heading Generation} - 
We use a follow-up question to generate the section headings while retaining the previous information, including references, subjects, title, and potentially, their explanations.
We instruct the model to generate between 6 and 10 headings. Although the LLM might not always adhere strictly to this range, it is only required to regenerate the headings if the initial count falls outside the range of 3 to 25.
We concatenate these headings and their indices with newline characters (\textbackslash n) to form the \textit{outline} for the following prompts.

\begin{table}[h!]
\small
\centering
\begin{tabular}{p{11.7cm}}
\textbf{Prompt-2, section heading generation (a follow-up question):}\\
Can you guess the outline of this paper? Just generate a list of first-level headings. \\
About 6-10 first-level headings are good! 
Just output the first-level headings, do not generate any other content.\\
No item number is required, each first-level heading begins with ``* ".
\end{tabular}
\end{table}

\paragraph{Step 3: Abstract Generation} - 
We use the following prompt to generate the abstract of the paper by summarizing the generated title and outline under the given subjects.

\begin{table}[h!]
\small
\centering
\begin{tabular}{p{11.7cm}}
\textbf{Prompt-3, abstract generation:}\\
You are an academic paper writing assistant in the subject \textit{``\{subject\}"}. \\
I am writing a survey paper titled with \textit{``\{title\}"}. \\
Here is my outline :
\textit{\{outline\}} \\
Can you write an abstract for me? \\
You may write only 1 paragraph with about 200-500 words, do not include more detailed headings, or any lists. \\
You should focus on the content of the abstract, do not repeat the word ``abstract" or including any other content.
\end{tabular}
\end{table}

\paragraph{Step 4: Selecting References for Each Section} - 
The full reference list is so extensive that using it in the subsequent generation steps would incur high API usage costs. 
Additionally, conditioning each section on the same group of references may lead to similarity across sections. 
Therefore, we select a subset of references for each section before generating the detailed structure and content.

Specifically, for each section title generated in step 2, we instruct the LLM to generate relevant references using the following prompt. 
The generated references will be used in the generation of subsection headings and contents for the specific section.

\begin{table}[h!]
\small
\centering
\begin{tabular}{p{11.7cm}}
\textbf{Prompt-4, reference selection:}\\
You are an academic paper writing assistant in the subject \textit{``\{subject\}"}. \\
I am writing a survey paper titled with \textit{``\{title\}"}. \\
Here is my outline : \textit{\{outline\}} \\
I will give you a list of references. Can you help me choose from these references that might be useful when I write this chapter the section \textit{``\{sec heading\}"}? \\
Here are the references: \textit{\{references\}} \\
The output should be in several lines. 
References that you think may be useful should be on one line each, beginning with ``*''.  \\
Please retain the square bracketed numbers (like [1], [20]) I give for each reference.
\end{tabular}
\end{table}

\paragraph{Step 5: Subsection Heading Generation} - 
We use prompt-5 to generate the subsection headings. 
The LLM leverages information from the title, outline, and the selected references to generate subsection headings for a given section. 
We then extend the outline generated in step 2 by incorporating the subsection headings for all sections, providing a comprehensive outline for content generation in the next step.

\begin{table}[h!]
\small
\centering
\begin{tabular}{p{11.7cm}}
\textbf{Prompt-5, subsection heading generation:}\\
You are an academic paper writing assistant in the subject \textit{``\{subject\}"}.  \\
I am writing a survey paper titled with \textit{``\{title\}"}. \\
Here is my outline: \textit{\{outline\}} \\
Here a list of reference papers: \textit{\{subset of references\}}\\
I am working on the section \textit{``\{sec heading\}"}.\\
Can you write the second-level headings for this section? 3-5 second-level headings are cool!\\
Just output the corresponding second-level headings, do not generate any other content.\\
No item number is required, each second-level heading begins with "* ".
\end{tabular}
\end{table}

\paragraph{Step 6: Content Generation} - 
Finally, detailed content is generated for each subsection using prompt-6. 
To encourage the content is coherent and aligned with the overall structure, we provide an outline consisting of all the section and subsection headings. We also supply the subset of relevant references generated in step-4 and encourage the model to cite them appropriately by an example. 

We observed that the prompt can become quite lengthy, causing the LLM to lose focus on the main task of generating subsection content. To address this, we emphasize the subsection title several times within the prompt to ensure the model remains focused.

\begin{table}[h!]
\small
\centering
\begin{tabular}{p{11.7cm}}
\textbf{Prompt-6, subsection content generation:}\\
You are an academic paper writing assistant in the subject \textit{``\{subject\}"}. \\
I am writing a survey paper titled with \textit{``\{title\}"}. \\
Here is my outline : \textit{\{outline\}}\\
Here a list of reference papers. I hope you will cite these references in the generated content. For example, [123] indicates a reference to the reference that begins with [123].
\textit{\{subset of references\}} \\
Can you write the content of \textit{``\{subsec heading\}"} part in the section \textit{``\{sec heading\}"}? 
You may write 3-5 paragraphs with about 1000 words, do not include more detailed headings. 
You should focus on the subsection \textit{``\{subsec heading\}"}, do not repeat the headings or including other content.
\end{tabular}
\end{table}

\section{Experiments}
\subsection{Dataset}

We use the dataset from the NLPCC 2024 \textit{Scientific Literature Survey Generation} evaluation task, provided by Kexin Technology \cite{tian2024overview}. 
The dataset comprises randomly crawled arXiv survey papers along with their references, containing 500 survey papers in the training set and 200 in the test set. 
Since the evaluation task did not designate specific instances for validation rather than training, we have not included a separate validation set in our statistics.

In the training set, the fields including \textit{title}, \textit{article id}, \textit{subject}, \textit{abstract}, \textit{content}, \textit{reference}, and \textit{reference content} are provided. 
While in the test set, only \textit{subject}, \textit{reference}, and \textit{reference content} are provided.
The statistics of the dataset is shown in Table.~\ref{tab:statistics}.

\begin{table}[t!]
\small
\centering
\setlength\tabcolsep{8pt}
\begin{tabular}{l|ccccc}
\toprule
 & \#paper & avg.subject & avg.ref & avg.ref\_content & avg.content \\
\midrule
Train & 500 &1.7&~98.5&21.1&72.5k\\
Test & 200 &1.9&126.8&27.2&--\\
\bottomrule
\end{tabular}
\caption{\label{tab:statistics}Dataset statistics, including the total number of papers, the average number of subjects, references, references with content, and the average content length, respectively.}
\end{table}

In this paper, we only use the training set for prompt design, instead of using them for parameter tuning or in-context learning. 
As discussed in \S\ref{sec:task}, because of the requirements in task guideline, we did not utilize the \textit{reference content} field provided in the test set, nor did we employ any retrieval-based methods to augment the reference titles.

\subsection{Implement Detail}

We implemented our system using the Qwen-long version of Alibaba's Tongyi Qianwen model \cite{bai2023qwen}, with a cost of 0.5 RMB per million tokens for input and 2 RMB per million tokens for output. 
For instances where the reference list is too long for Qwen-long to process (with the role of \textit{user}), we truncate the references, leaving only the first 80-100 references. 
Additionally, in cases where the reference list contains inappropriate content that Qwen-long refuses to process, we try to use references with odd or even indices to circumvent this issue. Generating all 200 surveys in the test set cost 20.86 RMB, approximately 0.10 RMB per paper. This low cost enhances the practicality of our method.

\subsection{Evaluation Metric}

To evaluate the quality of the generated literature survey, we use three kinds of metrics:

\textbf{ROUGE} \cite{lin2004rouge} is a recall-oriented reference-based metrics. Specifically, we employ ROUGE1, ROUGE2, and ROUGEL scores. 
We use the implement provided by Google\footnote{https://github.com/google-research/google-research/tree/master/rouge}.

\textbf{Soft Heading Recall} (S-H Recall for short) is used to evaluate the structure of the generated survey.

$$
\text{Sim}(t_i, t_j) = \cos \left( \text{embed}(t_i), \text{embed}(t_j) \right)
$$
$$
\text{card}(T) = \sum_{i=1}^{\left|T\right|} \frac{1}{\sum_{j=1}^{\left|T\right|} \text{Sim}(t_i, t_j)}
$$
$$
\text{card}(R \cap G) = \text{card}(R) + \text{card}(G) - \text{card}(R \cup G)
$$
$$
\text{soft heading recall} = \frac{\text{card}(R \cap G)}{\text{card}(R)}
$$
$T = \{t_1, t_2, t_3, \ldots, t_K\} $ represents a group of the chapter titles/heads in a generated/reference survey. 
$R$ and $G$ are the chapter titles of the generated and reference survey, respectively.
The bge-large-en-v1.5\footnote{https://huggingface.co/BAAI/bge-large-en-v1.5} model is used for text embedding.
This score encourages the similarity between generated and reference chapter titles while punishes the similarity of titles within the generated survey.

\textbf{Human} evaluation is also used to manually evaluate the following aspects:
\begin{itemize}
\item Fluent language with clear expression;
\item Logical article structure;
\item Ample, reliable, and accurate citations;
\item Consistency of content with the theme, staying on-topic;
\item Broad analytical scope.
\end{itemize}

The overall \textbf{score} is calculated as the average of the ROUGE score, the soft heading recall, and the human score. 
Note that the ROUGE score here is the average of ROUGE1, ROUGE2, and ROUGEL.

\subsection{Evaluation Result}

\begin{table}[t!]
\centering
\setlength\tabcolsep{4pt}
\begin{tabular}{l|cccc|c|c}
\toprule
& \scriptsize \textbf{S-H Recall} & \scriptsize \textbf{ROUGE1} & \scriptsize \textbf{ROUGE2} & \scriptsize \textbf{ROUGEL} & \scriptsize \textbf{Human} & \scriptsize \textbf{Score} \\
\midrule
Flying & \textbf{97.01} & 43.99 & 12.79 & 13.07 & \textbf{70.62} & \textbf{63.64} \\
CNKI-Research-Group & 90.21 & \textbf{45.95} & 14.07 & \textbf{14.16} & 68.48 & 61.14 \\
\textbf{ID (Ours)} & 95.84 & 43.53 & \textbf{14.68} & 12.91 & 63.78 & 61.11 \\
\tiny Literature Summarizer Wizards & 95.06 & 41.58 & 13.29 & 13.41 & 57.66 & 58.49 \\
BaseLine & 91.03 & 37.07 & 10.70 & 12.25 & 57.85 & 56.30 \\
Literature hunter & 94.03 & 37.07 & 11.54 & 12.35 & 54.45 & 56.27 \\
DUFL2024NLP & 87.48 & 26.71 & 7.81 & 9.54 & 52.22 & 51.46 \\
IDEA of NUDT & 84.59 & 27.26 & 5.93 & 9.95 & 35.97 & 44.98 \\
\bottomrule
\end{tabular}
\caption{Evaluation Results of Participating Teams\label{tab:evaluation_results}}
\end{table}

The official evaluation results is shown in Table~\ref{tab:evaluation_results}.
We can see that our system, \textbf{ID}, achieved the third place in the evaluation task, with an overall score only 0.03\% lower than the second-place team.

\begin{table}[th!]
\small
\centering
\resizebox{0.92\textwidth}{!}{
\begin{tabular}{p{11.7cm}}
\toprule
\textbf{Title} \\
\textbf{Reference:} On Interpretability of Artificial Neural Networks: A Survey \\
\textbf{Ours:} A Comprehensive Survey on Explainable Machine Learning and Artificial Intelligence: Methods, Applications, and Challenges\\
\midrule
\textbf{Outline} \\
\textbf{Reference:} \\
I.  INTRODUCTION \\
A. What Does Interpretability Mean? ~~~
B. Why Is Interpretability Difficult? ~~~
C. How to Build a Good Interpretation Method?  \\
II.  A SURVEY ON INTERPRETATION METHODS \\
A. Taxonomy Definition ~~~
B. Post-hoc Interpretability Analysis ~~~
C. Ad-hoc Interpretable Modeling \\
III.  INTERPRETABILITY IN MEDICINE \\ 
IV.  PERSPECTIVE 
V.  CONCLUSION 
VI.  ACKNOWLEDGEMENT\\
\textbf{Ours (partial):} \\
1. Introduction \\
1.1 Background and Importance of Explainable ML/AI ~~~
1.2 Historical Perspective and Evolution of XAI ~~~
1.3 Key Challenges and Objectives in Explainability\\
...\\
4. Saliency Maps and Visualization Techniques\\
5. Auditing and Fairness in Black-Box Models \\
...\\
15. Ethical Considerations and Regulations \\
16. Future Directions and Open Challenges\\
\midrule
\textbf{Abstract} \\
\textbf{Reference} : 
... In this paper, we propose a simple but comprehensive taxonomy for interpretability, systematically review recent studies in improving interpretability of neural networks, \textbf{describe applications of interpretability in medicine}... \\
\textbf{Ours:} 
... The paper explores various XAI techniques, including Case-Based Reasoning, Saliency Maps, and Shapley Values, while examining their applications and limitations. It also discusses the \textbf{use of Deep Learning in healthcare} and the interpretability aspects of Generative Adversarial Networks (GANs) and Recurrent Neural Networks (RNNs) in Natural Language Processing. ...
\\
\midrule
\textbf{Content \& Citation}\\
\textbf{Reference} : \\
Also, interpretability plays an important role in ethic use of deep learning techniques [57]. \textbf{To build patients’ trust in deep learning}, interpretability is needed to hold a deep learning system accountable [57]. \\
\textbf{Ours:} \\
Additionally, the ethical concerns surrounding data privacy, bias, and algorithmic accountability need to be addressed. Regulatory frameworks, such as those outlined in [57], must adapt to accommodate the integration of \textbf{AI technologies in healthcare while maintaining patient safety and privacy.}\\
\textbf{Citation information:} \\
$\left[57\right]$ J. R. Geis, et al., “Ethics of artificial intelligence in radiology: summary of the joint European and North American multisociety statement,” Canadian Association of Radiologists Journal, vol. 70, no. 4, pp. 329-34, 2019. \\
\bottomrule
\end{tabular}
}
\caption{A comparison between our generated survey and the reference paper \cite{fan2021interpretability}. \label{tab:case study}}
\end{table}

Additionally, our system demonstrates competitive performance across other metrics. Notably, in soft heading recall, our system achieves a score of 95.84\%, securing second place and trailing the leading team by just 1.17\%. 
This result highlights its effectiveness in generating high-quality headings. 
We attribute this success to our sequential generation design, which includes an individual heading generation step, providing a high-level perspective that enhances the ability of large language models to generate meaningful headings.

However, human evaluation remains a relatively weak aspect of our method. 
Without leveraging the content of the reference papers, the hallucination phenomenon of LLMs becomes significant. 
Consequently, our method cannot ensure the accuracy and reliability of the citations and analysis in the generated survey, thereby impairing our human evaluation score. 
In the future, we will incorporate the reference content into our framework to improve the factual accuracy and reliability of our outputs, aiming to generate more accurate and trustworthy literature surveys.

\subsection{Case Study}

An example of our generated survey paper and the corresponding reference one is shown in Table~\ref{tab:case study}. Our system successfully captures the subject of the paper, \textit{interpretability in neural networks}, and generates a relevant title.

For the outline generation, this example achieves a soft heading recall of 97.78\%. While the outline produced by our system is somewhat lengthy and too detailed, it successfully captures key aspects of the survey, such as \textit{explainable AI}, \textit{Saliency Maps}, and \textit{Black-Box Models}. 
The discussion on \textit{ethical considerations} and \textit{future directions} at the end of the survey is also well-founded. In the future, some specific instruction strategies can be designed to encourage the LLM to generate more concise outlines.

For the abstract and content (citation), we can see that our generated paper captures several common points with the reference one (\textbf{bold} in Table.~\ref{tab:case study}), resulting in appropriate abstracts and citations. 
However, some citations are still influenced by hallucinations, which could be mitigated by incorporating the detailed content of the reference papers.

\section{Conclusion}

In this paper, we presented a novel approach to generating scientific literature surveys with LLMs. We designed a series of instructions to generate well-structured and relevant content in a step-by-step manner. 
Our system with Qwen-long achieved third place in the NLPCC 2024 \textit{Scientific Literature Survey Generation} evaluation task, with an overall score only 0.03\% lower than the second-place team. 
However, challenges remain in ensuring the accuracy and reliability of citations and content due to the hallucination of LLMs and the absence of reference content. 
Future work will focus on addressing these limitations to generate more robust and trustworthy automatic literature survey.

\section*{Acknowledgements}

This work is supported by NSFC (62206070) and the Innovation Fund Project of the Engineering Research Center of Integration and Application of Digital Learning Technology, Ministry of Education (1221014). The datasets used in this paper is supported from Kexin Technology.

\end{document}